\documentclass[conference]{IEEEtran}
\IEEEoverridecommandlockouts
\usepackage{cite}
\usepackage{amsmath,amssymb,amsfonts}
\usepackage{algorithm}
\usepackage{graphicx}
\usepackage{textcomp}
\usepackage{xcolor}
\usepackage{subfigure}
\usepackage{algorithmic}
\usepackage{enumitem}
\def\BibTeX{{\rm B\kern-.05em{\sc i\kern-.025em b}\kern-.08em
    T\kern-.1667em\lower.7ex\hbox{E}\kern-.125emX}}
\begin{document}

\title{CCC: Color Classified Colorization
}

\author{\IEEEauthorblockN{ Mrityunjoy Gain, Avi Deb Raha, and Rameswar Debnath*,\textit{ Member, IEEE}}
	\IEEEauthorblockA{\textit{Computer Science and Engineering Discipline, Khulna University, Khulna 9208, Bangladesh}\\ 
	E-mail: \{gain1624, dev1611, rdebnath\}@cseku.ac.bd}
}

\maketitle

\begin{abstract}
Automatic colorization of gray images with objects of different colors and sizes is challenging due to inter- and intra-object color variation and the small area of the main objects due to extensive backgrounds. The learning process often favors dominant features, resulting in a biased model. Like the class imbalance problem, a weighted function imposing a higher weight on minority features can solve this feature imbalance problem. In this paper, we formulate the colorization problem into a multinomial classification problem and then apply a weighted function to classes.  We propose a set of formulas to transform color values into color classes and vice versa. Class optimization and balancing feature distribution are the keys for good performance. Class levels and feature distribution are fully data-driven. Observing class appearance on various extremely large-scale real-time images in practice, we propose 215 color classes for our colorization task. During training, we propose a class-weighted function based on true class appearance in each batch to ensure proper color saturation of individual objects. We establish a trade-off between major (mostly appearing) and minor classes (rarely appearing) to provide orthodox class prediction by eliminating major classes' dominance over minor classes. As we apply regularization to enhance the stability of the minor class, occasional minor noise may appear at the object's edges. We propose a novel object-selective color harmonization method empowered by the Segment Anything Model (SAM) to refine and enhance these edges. We propose a new color image evaluation metric, the Chromatic Number Ratio (CNR), to quantify the richness of color components. We compare our proposed model with state-of-the-art models using five different datasets: ADE, Celeba, COCO, Oxford 102 Flower, and ImageNet, in both qualitative and quantitative approaches. The experimental results show that our proposed model outstrips other models in visualization and CNR measurement criteria while maintaining satisfactory performance in regression (MSE, PSNR), similarity (SSIM, LPIPS, UIQI), and generative criteria (FID).
\end{abstract}


\section{Introduction} \label{sec:intro}
\begin{figure}[ht]
\centering
    \includegraphics[width=\linewidth, height = 3.5cm]{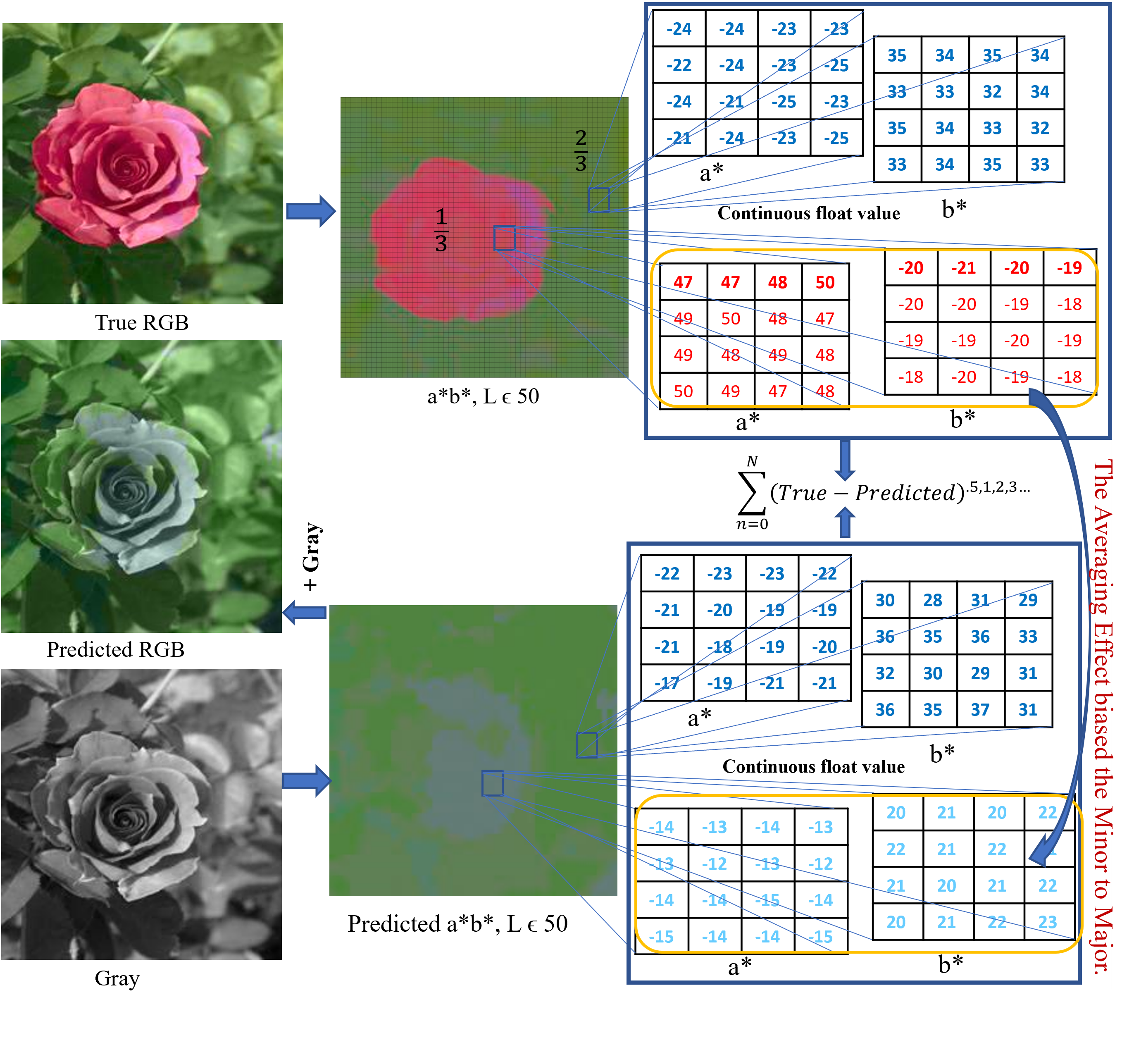}
    \caption{Imbalance feature distribution makes the regression task biased}
    \label{fig:subfig1}
\hfill
    \includegraphics[width=\linewidth, height = 3.5cm]{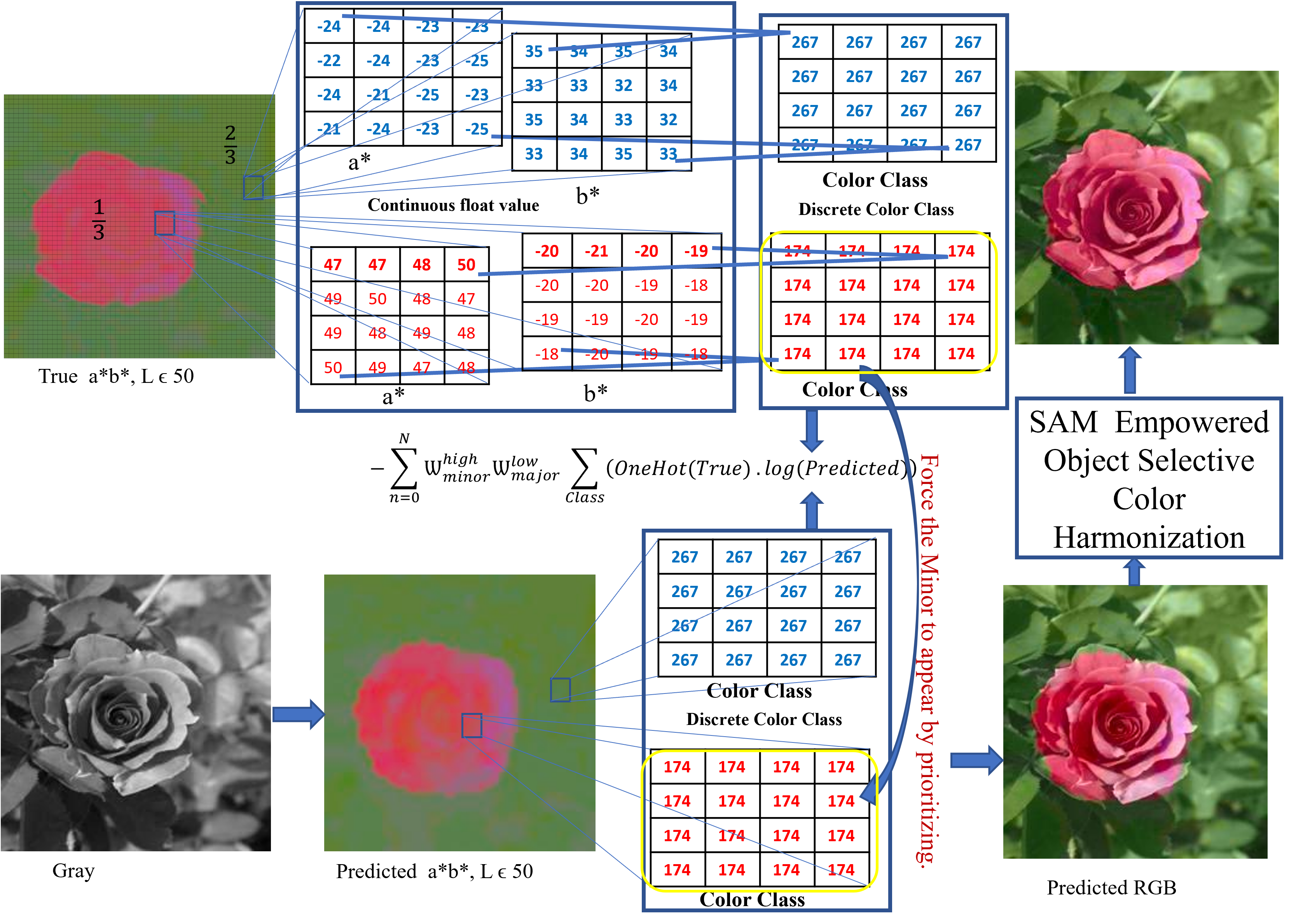}
    \caption{Special priority can effectively regulate the minor class}
    \label{fig:subfig2}
\caption{CCC can overcome feature imbalance in colorization by considering the regression task as a classification task and imposing higher weights on minor classes. }
\vspace{-5mm}
\end{figure}
Human vision perceives thousands of colors, making object identification easier. Color images are a popular way to express creativity and reminisce. Colorizing images from antiquity, medicine, industry, and astronomy helps convey their meanings. Color-coded subjects continue to captivate the public with remastered versions of vintage black-and-white movies, colored books, and online automatic colorization bots.\\  
Colorization is a process that assigns color components to grayscale images. It can be non-linear and ill-posed, allowing multiple colors in a single gray image. For example, a fruit's color can be light green, yellow, or red. Natural colorization aims to predict credible color distribution, not just the intensity values of a gray image. This process is not limited to the ground truth image color values.\\
Researchers have used various methods for image coloring, including user-guided \cite{Huang,Levin,Yatziv,Qu,Luan,Welsh,Ironi,Tai,Chia,Liu,Sousa,He,Zhang_tog,Charpiat,Gupta,Bugeau} and learning-based methods\cite{Wu1,Wu2,Wu3,Guo,Bahng,Liang,Zang,Larsson,Gain1, Gain2,An,Iizuka,Su,Dahl,Baldassarre,Zhang_eccv,Xia,Hesham,Ozbulak,Kong,Treneska}. Traditional user-guided methods require significant human interaction, leading to a decline in effectiveness. Learning-based strategies, which involve classical regression\cite{Gain1,Iizuka,Dahl,Baldassarre,Yun,Nguyen-Quynh}, object segmentation\cite{Su,Xia,Hesham,Kong}, generative approaches\cite{Wu1,Wu2,Wu3,Guo,Bahng,Liang,Zang,Ozbulak,Treneska}, and feature-balancing\cite{Larsson,Gain2,An,Zhang_eccv} techniques, are now more popular. Deep learning approaches, particularly in regression, are also gaining popularity for image colorization. These methods are easier to implement and require less human labor.
Deep Neural Networks (DNNs) learn representative features and hidden structural knowledge from data through training. The loss function generates feedback to refine the model's parameters, and networks adjust weights proportionally to the error. However, imbalanced class distributions can cause learning models to incorrectly classify minority class observations, making predicted class probabilities unreliable\cite{DWBL_12,Wallace_14}.\\
The colorization problem primarily involves feature distributions, with unbalanced distributions causing imbalances in the training process. Desaturated color components are more prevalent in training images, impacting the performance of saturated color components. This bias can cause smaller objects' hues to merge with the background, making learning tasks biased. Handling feature imbalance is essential for achieving the desired learning outcomes. \\
Class imbalance is often resolved by resampling the dataset or using weighted functions to increase minority class weight. In training, features of a sample determine gradient directions on the loss function. The sample's spatial resolution determines the input dimension of a learning model, while the output dimension of colorization models is the same. Defining rules to transform feature values into class values can solve feature imbalance problems. 

The study proposes a method to transform continuous color values into discrete color classes and vice versa to predict a distribution of possible colors for each pixel. The study revealed that 215 of the 400 color classes are predominantly present. So, we reduced the class to 215 as redundancy reduced the classification accuracy. To address the class imbalance issue, we determine the class weights by analyzing the true class of each batch during training, assigning higher weights to rarely-appearing classes. This adjustment aims to alleviate desaturation and biases towards predominant features. We propose a SAM-empowered\cite{SAM} object-selective color harmonization method to refine and polish the edge more.

Our proposed model formulated regression problem into classification problem based on the work CIC\cite{Zhang_eccv}. The CIC\cite{Zhang_eccv} defines classes that are static. But self-supervision majorly depends on data pattern and variety. So We formulate our problem in a data-driven manner. Moreover, we embedded SAM that improves results because it eradicates color bleeding and ensures object-selective color harmonization for model failure cases. We also describe with experiments the impact of classes with their appearance in the image ($>$500).
The basic works improvement of CCC over the work CIC\cite{Zhang_eccv} are illustrated in Tab. \ref{table1}:
\begin{table}[h]
    \centering
        \vspace{-2mm}
        \caption{Differences between the CIC\cite{Zhang_eccv} and our proposed CCC}
        \vspace{-2mm}
    \begin{tabular}{c c c}
         Contents & CIC\cite{Zhang_eccv} & CCC \\
        Formula for color class conversion and vice versa & No & Yes  \\
        Data driven class points optimization & No & Yes  \\
        Task generalization(Adaptibility on \\ similar feature imbalance problem) & No & Yes  \\
         Data driven class weight formulation & No & Yes  \\
         Segmentation based edge refinement & No & Yes \\
         Chromatic diversity evaluation metric & No & Yes
    \end{tabular}
    \label{table1}
\end{table}\\
The essence of our method is shown in Fig. \ref{sec:intro}. The following are the contributions to this work:
\begin{enumerate}
 
  \item We propose a set of formulas to transform continuous double-channel color values into discrete single-channel color classes and vice versa. Any feature imbalance regression problem can be configured to a classification problem using these baseline formulas.

  \item We optimize class levels of the colorization problem by analyzing numerous different images.
 
  \item We propose a class re-weighting formula for graving high gradient from misclassified low appeared or rare classes to ensure a balance contribution of all classes in the loss. This removes feature biases as well as desaturation along with over-saturation from the color distribution and ensures orthodox prediction.

   \item We proposed a novel object-selective color harmonization method empowered by the Segment Anything Model (SAM) to make the edge more refined and polished.
   
  \item We propose a new color image evaluation metric, Chromatic Number Ratio (CNR), which quantifies the richness of color classes in generated images compared to ground truth images, providing a comprehensive measure of the color spectrum.
 
  \item We present an abundance of quantitative and qualitative results demonstrating that our method significantly outperforms extant state-of-the-art baselines and produces reasonable results.

\end{enumerate}
The rest of the paper is structured as follows: Sec. \ref{related work} reviews the relevant literature; Sec. \ref{sec:methodology}, the entire CCC, including problem formulation and solution approach; Sec. \ref{sec:sam}, the SAM-empowered color harmonization; Sec. \ref{experiment} the experimental outcomes and a comparative analysis with other cutting-edge techniques; and Sec. \ref{conclusion}, the conclusion. 

\section{Related Litarature}\label{related work}

Image colorization mainly falls into two categories: user-guided and learning-based colorization.
\subsection{User Guided Colorization}
The user-guided colorization mainly falls into two categories: scribble-based and example-based.\\
\textbf{Scribble-based colorization} The scribble-based colorization technique uses user input to fill in missing or incomplete sections of an image. Techniques include optimizing color propagation\cite{Levin}, combining non-iterative techniques with adaptive edge extraction\cite{Huang}, introducing color blending\cite{Yatziv}, propagating color effectively\cite{Qu} in pattern-continuous and intensity-continuous regions\cite{Luan}, and incorporating U-Net structures\cite{Zhang_tog}. \\
\textbf{Example-based colorization} The example-based colorization minimizes user effort in grayscale image transmission, including global color statistics\cite{Welsh}, segmented region-level approaches\cite{Ironi,Tai,Charpiat}, super-pixel-level\cite{Gupta,Chia}, and pixel-level methods\cite{Liu,Bugeau}. However, manual similarity metrics can be prone to error in scenarios with significant variations in intensity and content\cite{Sousa,He}.
 
\subsection{Learning Based Colorization}
Learning-based colorization is a machine-learning technique that automatically applies color to grayscale or black-and-white photographs using CNNs trained on large datasets. The main challenge is feature balancing for focused objects and backgrounds.\\
\textbf{Basic Regression Based Colorization:}
Colorization involves using conventional CNN or specialized architectures like InceptionNet, VGGNet, ResNet, and DenseNet to estimate color channels from grayscale images. Gradient is calculated using regression loss function by automated methods, such as\cite{Dahl, Huang, Baldassarre, Iizuka}, encoder-decoder based colorization models\cite{Nguyen-Quynh,Gain1}, iColoriT\cite{Yun} etc. \\
\textbf{Object Segmentation Based Colorization}
Various colorization models that segment objects within an image, learn color assignment segment-wise or object-wise, and assign colors to segments using techniques like spatial connections or global color coherence. These models have been developed using various techniques, including semantic segmentation\cite{Su,Hesham,Xia}, adversarial edge-aware models\cite{Kong}, and point annotations.\\
\textbf{GAN Based Colorization}
GAN image colorization models combine discriminator and generator networks to produce realistic, aesthetically pleasing colorized photographs. These models use semantic information\cite{Wu1}, CapsNet\cite{Ozbulak}, GAN encoders\cite{Wu2}, and other techniques\cite{Bahng,Guo} to improve colorization results. Examples include creating ethnic costumes, using GAN encoders for colorization, and using GAN for colorizing medical images\cite{Liang,Treneska,Wu3,Zang}. Techniques like transfer learning and deep convolution GAN have been developed for various applications.\\
\textbf{Feature Balancing for Colorization}
Zhang et al.\cite{Zhang_eccv} proposed an automatic colorization using CNN, classifying intensity into predetermined color levels and assigning corresponding colors based on classified class levels. An et al.\cite{An} used a VGG-16 CNN model and color rebalancing technique to solve feature imbalance problems. Larsson et al.\cite{Larsson} used unbalanced loss of classification, and Gain et al.\cite{Gain2} proposed a deep localized network for image colorization.

\section{Color Classified Colorization}\label{sec:methodology}
\subsection{Color Space}
Conventional RGB is the most commonly used color space, consisting of Red, Green, and Blue. However, its inability to distinguish between color and content information renders it inappropriate for color manipulation tasks involving colorization. CIE LAB\cite{CIE} is a suitable choice, as it separates color information from context information, allowing for manipulation while keeping context information unchanged. In La*b* (LAB) space, L denotes the brightness or luminosity of the picture, with intensities falling between $[0, 100]$. As $L$ increases, colors become brighter. The a* and b* channels correspondingly represent the image's proportion of red-green and yellow-blue tones, with red-yellow represented by a positive value and green-blue by a significant negative value, often falling between $[-128, 127]$.
\begin{figure}[!t]
\centering
\includegraphics[width=.98\columnwidth]{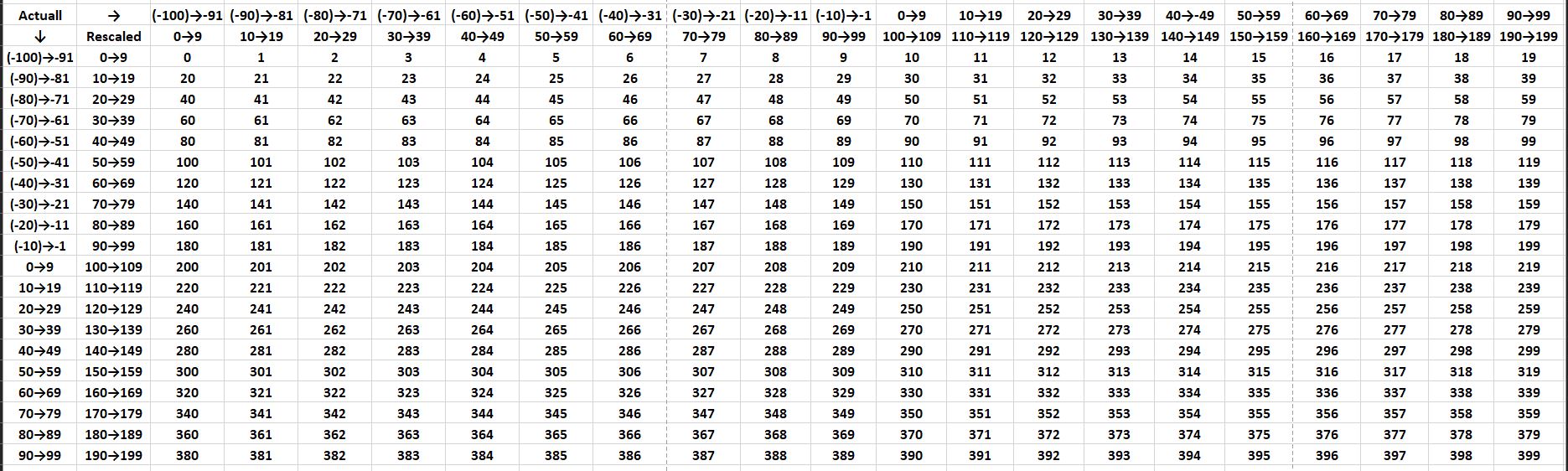}
 \caption{Color class conversion}
\label{ab_to_class}
\vspace{-2mm}
\end{figure}
\vspace{-2mm}
\subsection{Problem Definition}
The colorization problem is considered to predict color channels from a given gray channel. The Lightness (L) channel of La*b* color space can be mapped into the gray channel(intensity) and vice versa\cite{CIE}. Furthermore, RGB can be mapped into LAB and vice versa. The task can be defined as follows in Eq. \ref{problem_formulation_1}, \ref{problem_formulation_2}, \ref{problem_formulation_3}, and \ref{problem_formulation_4}.
\begin{equation}
\label{problem_formulation_1}
\small
\mathcal{X}_{ab} = f(\mathcal{X}_L) 
\vspace{-2mm}
\end{equation}
\begin{equation}
\label{problem_formulation_2}
\small
Distance_{min}(\mathcal{Y}_{ab}, \mathcal{X}_{ab})
\end{equation}
\begin{equation}
\label{problem_formulation_3}
\small
\mathcal{X}_{Lab} = concat(\mathcal{X}_L, \mathcal{X}_{ab})
\end{equation}
\begin{equation}
\label{problem_formulation_4}
\small
\mathcal{X}_L \in \mathbb{R}^{H\times W \times 1} ,
\mathcal{X}_{ab} \in  \mathbb{R}^{H\times W \times 2},
\mathcal{Y}_{ab} \in  \mathbb{R}^{H\times W \times 2}
\end{equation}
 where $\mathcal{X}_L$ is the lightness channel, $\mathcal{X}_{ab}$ is the predicted color channel, $\mathcal{X}_{Lab}$ is the predicted color image, $\mathcal{Y}_{ab}$ is the ground truth color channel, $f(.)$ is the mapping function achieved by deep learning, $Distance_{min}(.)$ is the objective function(can be any loss function) by which the optimizer makes the learning efficient, $\mathbb{R}$ is the total image component, $\mathbb{H}$ and $\mathbb{W}$ are the image dimension.

Theoretically, the values of the a* and b* channels are continuous within  [-128, 127]. Therefore, the prediction is considered a regression problem. That's why the $Distance_{min}(.)$ of Eq. \ref{problem_formulation_2} naturally can be either L1 loss or L2 loss or Huber loss or Log-cosh loss or similar regression loss shown in Eq. \ref{l1}, \ref{l2}, \ref{huber}, \ref{log-cosh}.
\begin{equation}
\label{l1}
\small
L_1(\mathcal{Y}_{ab}, \mathcal{X}_{ab}) = \frac{1}{N}\sum_{N}|\mathcal{Y}_{ab} - \mathcal{X}_{ab}|
\vspace{-2mm}
\end{equation}
\begin{equation}
\label{l2}
\small
L_2(\mathcal{Y}_{ab}, \mathcal{X}_{ab}) = \frac{1}{N}\sum_{N}(\mathcal{Y}_{ab} - \mathcal{X}_{ab})^2
\end{equation}
\begin{equation}
\label{huber}
\small
L_{\delta}=
    \left\{\begin{matrix}
        \frac{1}{N}\sum\limits_{N}\frac{1}{2}(\mathcal{Y}_{ab}-\mathcal{X}_{ab})^{2},&|\mathcal{Y}_{ab}-\mathcal{X}_{ab}|<\delta\\
        \frac{1}{N}\sum\limits_N\delta ((\mathcal{Y}_{ab} - \mathcal{X}_{ab}) - \frac1 2 \delta), & otherwise
    \end{matrix}\right.
\end{equation}
\begin{equation}
\label{log-cosh}
\small
Log-Cosh(\mathcal{Y}_{ab}, \mathcal{X}_{ab}) =  \frac{1}{N}\sum_{N}log(cosh(\mathcal{Y}_{ab} - \mathcal{X}_{ab}))
\end{equation}
Background colors like clouds, soil, pavement, and walls dominate real-time images, leading to an imbalanced distribution of features.  Handling feature imbalance is crucial because the smaller subsets of features are the feature of interest for the learning task. The ambiguity and multimodality of the colorization problem make the above loss functions vulnerable. The mean of the set is the most effective method to solve the loss, as the averaging error effect favors color values predominantly covered in the ground truth image. In an imbalanced feature distribution, the training process is biased towards larger feature subsets, resulting in the colors of smaller objects disappearing from the resulting models. The distribution of a*b* values is skewed towards desaturated values, causing the color of minuscule objects to disappear.
\begin{figure}[!t]
\centering
\includegraphics[width=0.95\columnwidth, height=2.5cm]{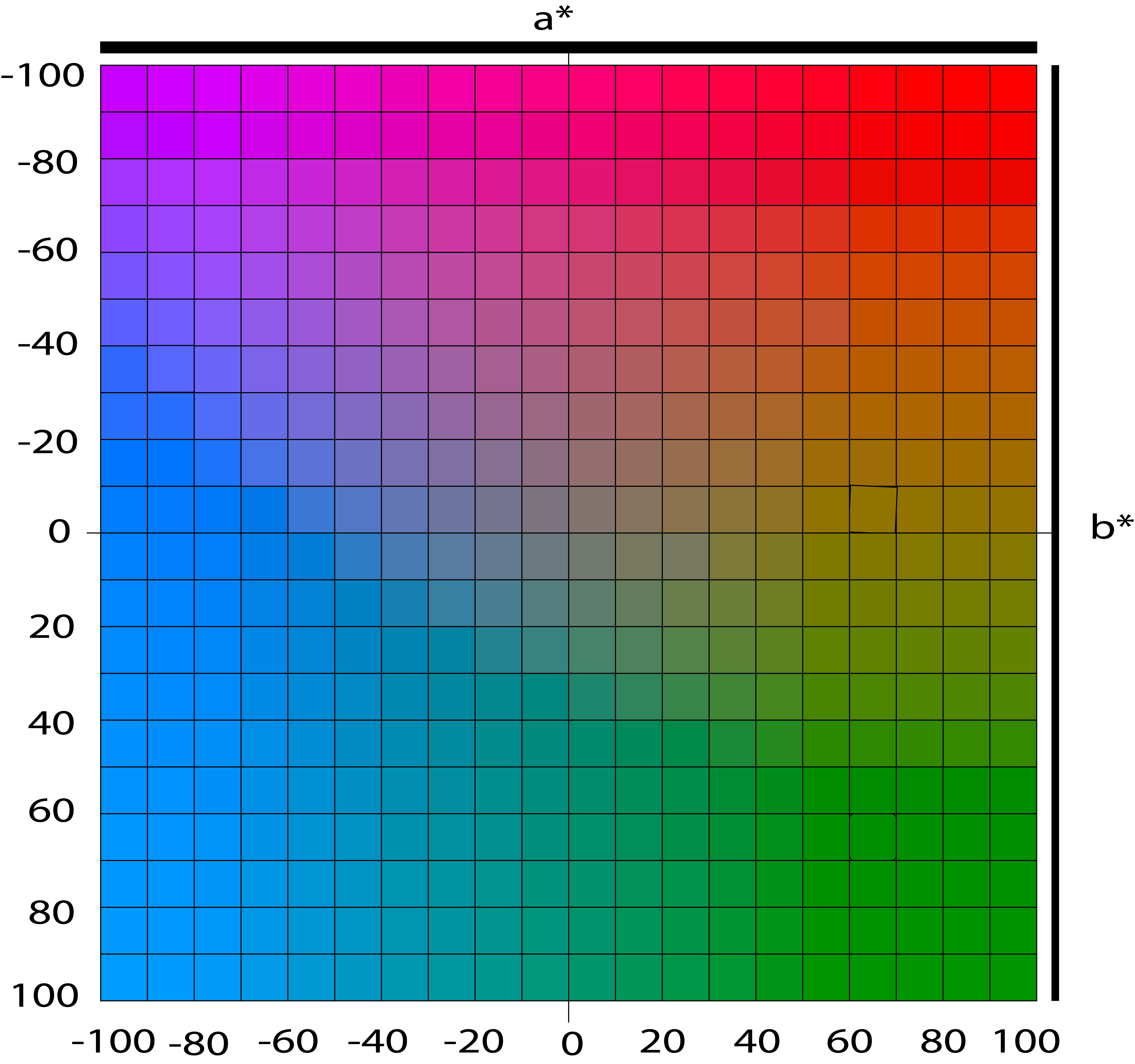}
\caption{Color class to visual color conversion}
\label{fig:class_to_ab}
\vspace{-4mm}
\end{figure}
\subsection{Solution Approach}
\begin{figure}[!t]
\centering
\includegraphics[width=0.95\columnwidth, height=2.5cm]{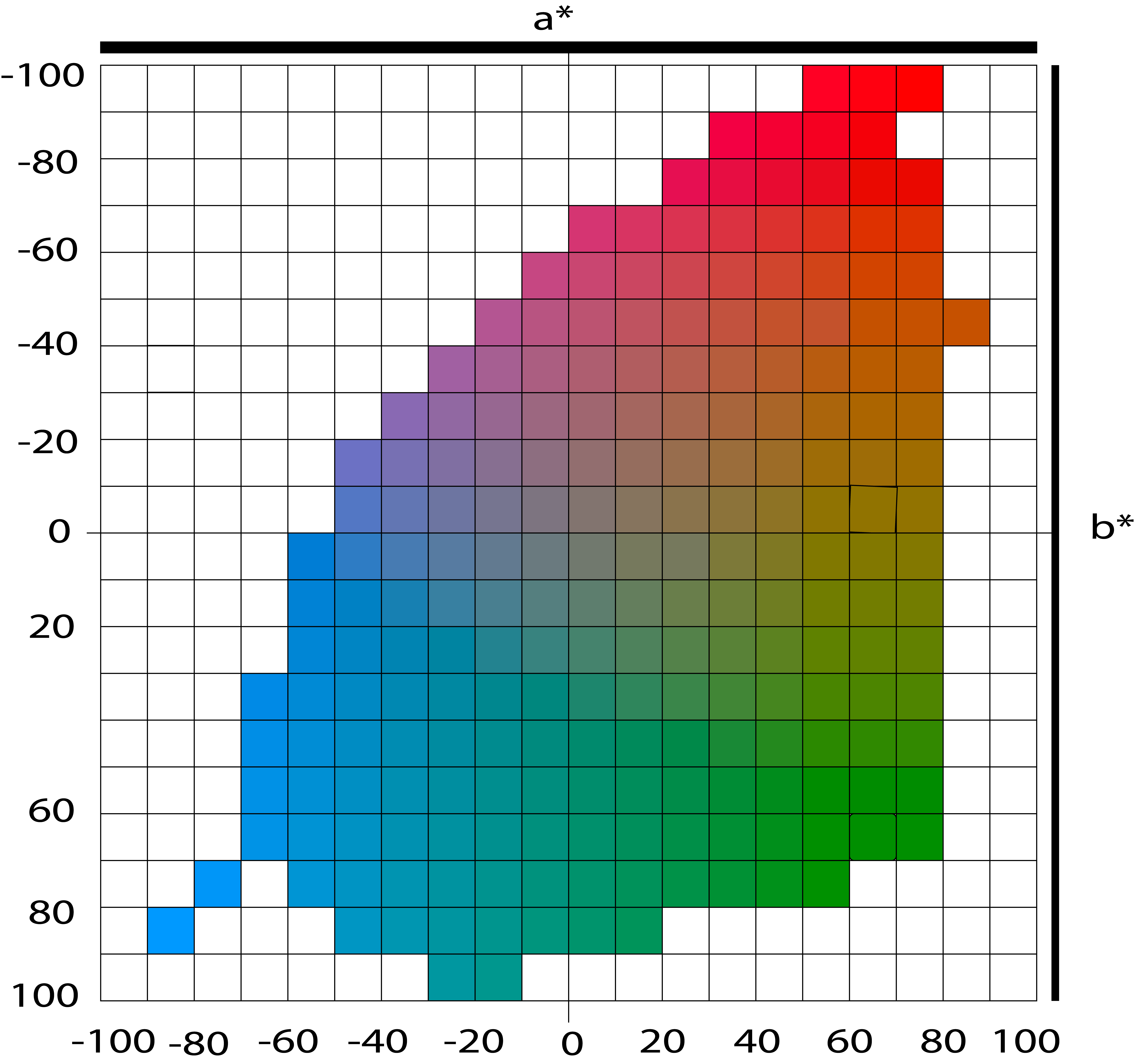}
\caption{Visualization of Real-time appeared of Color classes}
\label{ab_reduction}
\vspace{-2mm}
\end{figure}

\textbf{Continuous Color Range to Discrete Color Classes} \label{Continuous Color Range to Discrete Color Classes}
The a* and b* color channels are continuous within the $[-128, 127]$ range. Each a*b* pair with a lightness value $L$ forms an RGB color pixel. We can get an a*b* pair from a*b* color space, a 2-D space, where a* is one direction and b* is another. For a fixed $L$, a small change in the a*b* pair has no psychovisual effect. Because human perception of the information in an image normally does not involve quantitative analysis of every pixel value in the image. Colorization is a regression problem where the regression model predicts the continuous quantities of a* and b* for a given $L$. Taking advantage of the psycho-visual nature of humans, the colorization problem can be represented as a classification problem where the learning model predicts a discrete class level for an a*b* pair. To formulate the problem, the a*b* color space is divided into bins of a fixed grid size, and each bin is assigned a discrete class level. The formula is given below in Eq. \ref{color_to_class}.
\begin{equation}
\small
\label{color_to_class}
\mathcal{C} = \Big(\frac{b^*_i+\beta}{\alpha}\Big)\cdot \Delta + \frac{a^*_i+\beta}{\alpha}, \forall i \in N
\end{equation}
where a* and b* are the continuous color channels, $\mathcal{C}$ is the discrete color class, $\alpha^2$ is the area of a bin, $\beta$ is a shifting constant that shifts a*b* color values into the positive quadrant, $\Delta$ is the number of grids in each a* or b* color channel, $N$ is the total number of pixels.\\
\textbf{Color Class to Visual Color Mapping} 
We need to extract a*b* pairs from the predicted color classes, $\mathcal{C}$s, generated by the learning model for color image generation. Each bin is assigned by a fixed color class level $\mathcal{C}$ driven by $a^*$ and $b^*$.  The formulas are given below in Eq. \ref{class_to_color_a}, \ref{class_to_color_b}, which is the reverse of Eq. \ref{color_to_class}. 
\begin{equation}
\small
\label{class_to_color_a}
a^{*\prime} = [(\mathcal{C}_i\mod\delta)\cdot\alpha]-\beta +\frac{\alpha}{2} ,\forall i \in N
\end{equation}
\begin{equation}
\small
\label{class_to_color_b}
b^{*\prime} = [(\mathcal{C}_i\div\delta)\times\alpha]-\beta +\frac{\alpha}{2}, \forall i \in N
\end{equation}
According to the above equations, the maximum loss for each a* or b* value is $\frac{\alpha}{2} -1$. The higher value of $\alpha$ reduces the number of classes but makes the representation lossy as a large continuous range is converted to a single class. However, handling the problem with the lower class is easy. The lower value of $\alpha$ increases the number of classes. In the colorization problem, more classes make the prediction less precise. It is important to adjust the number of class levels for $a*b*$ color space so that modified $a^*b^*$ can describe the image's color nature.  \\
\textbf{Color Class reduction Based on Practical Appearance}
\begin{figure}[!t]
\centering
\includegraphics[width=0.98\columnwidth]{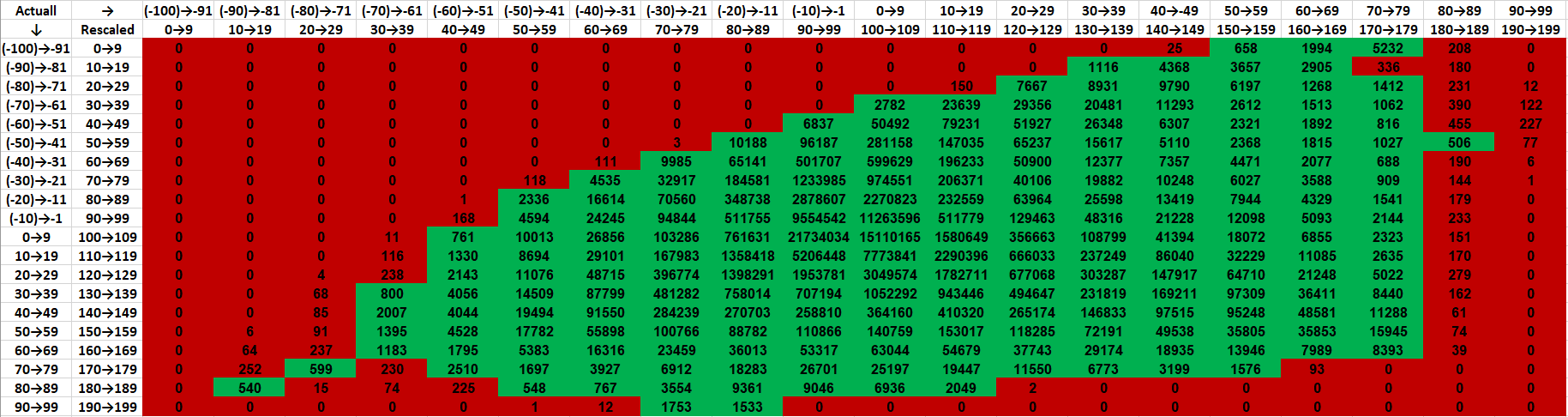}
\caption{Real-time appearance of Color classes}
\label{practical_color_class}
\vspace{-4mm}
\end{figure}
The a* and b* color values are continuous within $[-128, 127]$ in the a*b* color channel. But in practice, the range is found within $[-100, 99]$. We first transform the continuous $[-100, 99]$ ranged a*b* color channel to a single plane of 400 color classes by taking $\alpha = 10$, $\beta = 100$ and $\Delta = 20$ in Equation \ref{color_to_class}. A 2D grid of bins as $20\times20$ single plane array is then formed where horizontal axes indicate a* and vertical axes indicate b* color information. Each coordinate is assigned a class value. The class matrix is shown in Fig. \ref{ab_to_class}. Fig. \ref{fig:class_to_ab} shows the proposed color class to visual color. Fig. \ref{fig:class_to_ab} shows that the whole image is a smooth representation of different colors. The colors of the nearest bins or blocks are almost similar. Color changes gradually, block by block.  \\
The study focuses on the colorization of images using 400 color classes from the Place365 Validation dataset. We extracted the classes of 35040 images. The images were downsampled to $56\times56$ to reduce class samples, resulting in 109885440 class samples with 400 class levels. Each color pixel represents a color class sample with a specific color level. A class level is considered for training samples with a minimum of 500 (0.000455\%) class samples. Class samples under 500 with a specific class level are mapped to their nearest-neighbor present class levels using fixed centroid $k$-means clustering shown in Eq. \ref{kmeans}. The final color bin contains 215 color classes within 400 color classes with more than 500 pixels Which is shown in Fig. \ref{practical_color_class} and their visual in Fig. \ref{ab_reduction}. Class optimization is a major issue for the colorization model, as less class may make the model more error-free but may make some color visuals outside the bin. To keep rare color values in the predicted distribution, these visuals must be active in the training process.

\begin{equation}
\vspace{-2mm}
\small
\label{kmeans}
 kmeans(\mathcal{C},\mu) = arg min\sum_{i=1}^k\sum_{\mathcal{C}}||c- \mu_i||^2
\end{equation}
where $\mathcal{C}$ is the input color class vector, $\mu$ is the approved color classes for training, and $k$ is the number of color classes (215). We can define $\mu$ as the fixed value centroid. The iteration will happen a single time, and the centroid value will be unchanged.\\

\begin{figure*}[!t]
\centering
\includegraphics[width=\textwidth, height=5cm]{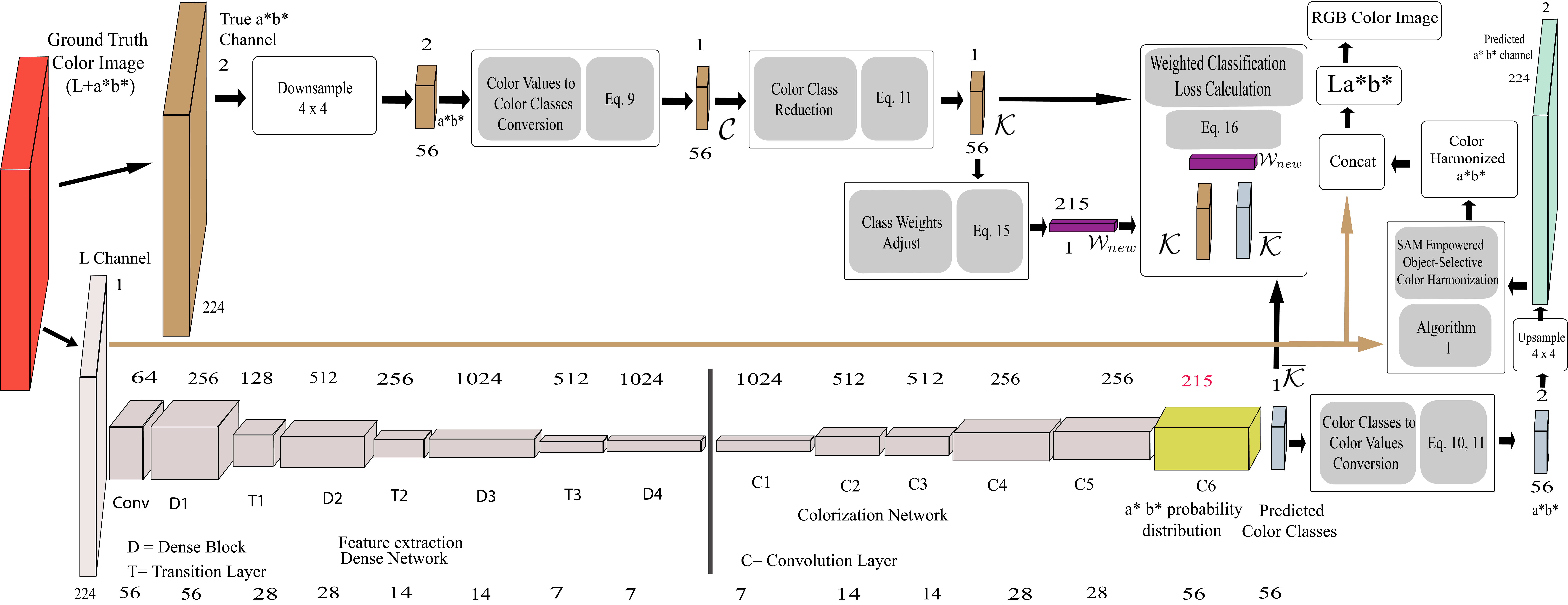}
\caption{Color classified Colorization}
\label{network}
\vspace{-4mm}
\end{figure*}

\textbf{Network Architecture}
We build our model based on an encoder-decoder architecture. We use DenseNet\cite{DenseNet} for the encoder part of our feature extractor. The DenseNet is a high-level feature extractor suitable for good color value generation. For the decoder part, we use conventional CNN. The Network architecture of our proposed method is shown in Fig. \ref{network}.
\begin{itemize}
\item {Feature Extraction}
DenseNet's robust connections minimize gradient vanishing and semantic information loss during feature extraction. It concatenates output from each layer, adapts to grayscale input by changing the first convolutional layer, and discards the final linear layer to build a $\frac{H}{32}\times\frac{W}{32}\times 1024$ feature representation.
\item {Colorization Network}
The network employs several convolutional and up-sampling layers after receiving an input of a $\frac{H}{32}\times\frac{W}{32}\times1024$ feature representation. The fundamental nearest-neighbor method is what we employ for up-sampling. The $56\times 56\times 215$ color class distribution is what the network outputs.
\end{itemize}
\textbf{Loss Calculation}
Colorization is generally considered a regression problem as the color values are continuous. But we transform the continuous color values into the discrete color classes. So, we consider the problem a classification problem and use cross-entropy loss instead of MSE or other regression loss. The loss function is shown in Eq. \ref{crossentropy}.
\begin{equation}
\small
\label{crossentropy}
Loss_{CE} =- \sum_{H,W}\mathcal{W}_c\sum_{\mathcal{C}} \mathcal{K} . log(\overline{\mathcal{K}}).
\end{equation}
Where $H$ and $W$ are the height and width of output $\mathcal{K}$ distribution. $\kappa$ is the true color class and $\overline{\mathcal{K}}$ is the estimated color class. The $\mathcal{W}_c$ is the weights vector of color classes. The $\mathcal{W}_c$ is defined as follows in Eq. \ref{normal_weights}.
\begin{equation}
\small
\label{normal_weights}
\mathcal{W}_c =  \Big(\frac{1}{n_\mathcal{C}}\Big), \forall_c \in \mathcal{C}
\end{equation}
where $n_C = 215 = $ the number of color classes.\\
\textbf{Class Confusion Based Weights Adjustment}
In realistic images, not all color classes are represented equally. Grayish visual color classes are found in a much larger proportion than bright color classes due to the large background areas. In the categorical cross-entropy loss, each true class gets $\frac{1}{N}$ weight during loss calculation, which is shown in Eq. \ref{normal_weights}. As the minor color classes are far smaller in the count values, the gradients disappear gradually iteration by iteration. To keep the rarely appearing color classes, we increase the weights of the rarely appearing color classes more than the mostly appearing color classes. However, this process increases the global loss. Therefore, the weights must be trade-offs to ensure plausible colors and a minimum loss. To trade off the weights, we proposed a new formula, which is given in Eq. \ref{weights_trade_off}.
\begin{equation}
\small
\label{weights_trade_off}
\mathcal{W}_{new} = \Bigg( \frac{\underset{c\in \mathcal{C}}{\max}(N_c)}{N_c \cdot \Upsilon + \underset{c\in \mathcal{C}}{\max}(N_c) \cdot \Phi}\Bigg), \forall c \in \mathcal{C}
\vspace{-1mm}
\end{equation}
where $\mathcal{C}$ is the color classes of a particular batch, ${\max}(N_c)$ is the maximum appearance value of a class, $N_c$ is the appearance value of class $c$, $\mathcal{W}_{new}$ is the new weights matrix of the particular batch, $\Upsilon$ and $\Phi$ is the trade-off factor where $\Upsilon$ can ranges (0,1] and $\Phi = \frac{1}{n_C}$ .\\
We initially normalize weights by dividing the count of the maximum appeared class in a batch by the total count of each 215 classes individually, ensuring the weight of the maximum class is set to 1 and proportionally up-scaling others. However, this approach leads to a significant increase in the weight of classes appearing very infrequently. To strike a balance, we introduce a trade-off mechanism. This involves adding a term, (${\max}(N_c) \cdot \Phi$), to the product of the individual class count ($N_c$) and a trade-off factor ($\Upsilon$). This supplementary term helps control the influence of rare class occurrences, providing a more nuanced and balanced approach to class weight determination.
Therefore, the loss function is now modified, as shown in Eq. \ref{crossentropy_new}.
\begin{equation}
\small
\label{crossentropy_new}
Loss_{CBCE} =- \sum_{H,W}\mathcal{W}_{new}\sum_{C} \mathcal{K} . log(\overline{\mathcal{K}}).
\vspace{-1mm}
\end{equation}
\textbf{Color Class Estimation}
The network outputs $H \times W \times C \times batch $ tensor. Using a softmax probability distribution, we extract $H \times W \times 1 \times batch$ class representation.
\begin{equation}
\small
\label{softmax}
\overline{\mathcal{K}} =\sigma(\mathcal{K}) = \frac{e^{\mathcal{K}_{i}}}{\sum_{j=1}^k{e^{\mathcal{K}_{j}}}}
\vspace{-1mm}
\end{equation}

\begin{algorithm}
\caption{SAM Empowered Object-Selective Color Harmonization}
\label{alg:segmentation}
\begin{algorithmic}[]
    \renewcommand{\algorithmicrequire}{\textbf{Input:}}
    \renewcommand{\algorithmicensure}{\textbf{Output:}}
    \REQUIRE Input gray image, Predicted $a^*b^*$
    \ENSURE Edge harmonized $a^*b^*$
    \STATE Extract segments ($\mathcal{S}$) from the gray image using SAM
    \FOR{each segment $s$ in $\mathcal{S}$}
        \IF{the number of pixels in $s > \Psi$}
            \STATE Extract $\mathcal{S}^a$ from $a^*$ using coordinates of $s$
            \STATE Calculate mode value ($\mathcal{M}_a$) of $\mathcal{S}^a$
            \FOR{each pixel ($\mathcal{P}$) in $\mathcal{S}^a$}
                \IF{$|\mathcal{P} - \mathcal{M}_a| > \delta_a$}
                    \STATE Replace the value of $\mathcal{P}$ with $\mathcal{M}_a$
                \ENDIF
            \ENDFOR
            \STATE Extract $\mathcal{S}^b$ from $b^*$ using coordinates of $s$
            \STATE Calculate mode value ($\mathcal{M}_b$) of $\mathcal{S}^b$
            \FOR{each pixel ($\mathcal{P}$) in $\mathcal{S}^b$}
                \IF{$|\mathcal{P} - \mathcal{M}_b| > \delta_b$}
                    \STATE Replace the value of $\mathcal{P}$ with $\mathcal{M}_b$
                \ENDIF
            \ENDFOR
        \ENDIF
    \ENDFOR
    \RETURN Edge harmonized $a^*b^*$
\end{algorithmic}
\end{algorithm}

\subsection{Chromatic Diversity}
We propose a novel color image evaluation metric named Chromatic Number Ratio (CNR). The CNR quantifies the richness of color classes within the generated images compared to the ground truth images. It offers a comprehensive measure of the spectrum of colors in the generated images, enhancing our understanding of color diversity. The metric is shown in Eq. \ref{color_diversity}.
\begin{equation}
\label{color_diversity}
\small
CNR =\frac{\sum\limits_{i=0}^{m-1} \sum\limits_{j=0}^{n-1} \left(1 - \sum\limits_{k=0}^{i-1} \sum\limits_{l=0}^{n-1} [\mathcal{P}_{i,j} = \mathcal{P}_{k,l}]\right)}{\sum\limits_{i=0}^{m-1} \sum\limits_{j=0}^{n-1} \left(1 - \sum\limits_{k=0}^{i-1} \sum\limits_{l=0}^{n-1} [\mathcal{G}_{i,j} = \mathcal{G}_{k,l}]\right)} 
\vspace{-1mm}
\end{equation}
where, $\mathcal{P}_{i,j}$ and $\mathcal{G}_{i,j}$ is the color class value at row $i$ and column $j$ of the generated color image $\mathcal{P}$ and Ground truth image $\mathcal{G}$. $m$ and $n$ are the image's dimensions in the color class space. The outer summation iterates through all rows ($i$) and columns ($j$) of the image in color class space. The inner summation compares each pixel ($\mathcal{P}_{i,j}$ / $\mathcal{G}_{i,j}$) with all previous pixels in the image to check for uniqueness. $[\mathcal{P}_{i,j} = \mathcal{P}_{k,l}]$ and $[\mathcal{P}_{i,j} = \mathcal{P}_{k,l}]$ is an indicator function that returns 1 if the condition is true (if pixel values are equal) and 0 if it's false.

Through the CNR, we have tried to show how many different color components are picked in the color images. We initially set color classes to 400. But in the training set, majorly appeared color classes dominate the minor classes. That's why the model overlooks the minor class in the prediction. We aim to ensure the minor classes are also in the predicted distribution while maintaining the other measurement criteria satisfactory. The CNR value 1 indicates the number of different color classes is the same. The predicted distribution can also pick more minor color classes making the visual more plausible and the CNR value greater than 1.
\begin{figure*}[!ht]
    \centering
        \includegraphics[width=0.085\linewidth]{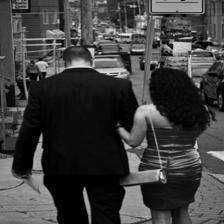}
        \includegraphics[width=0.085\linewidth]{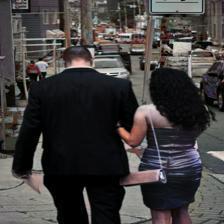}
        \includegraphics[width=0.085\linewidth]{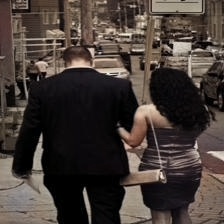}
        \includegraphics[width=0.085\linewidth]{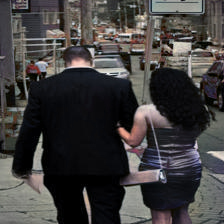}
        \includegraphics[width=0.085\linewidth]{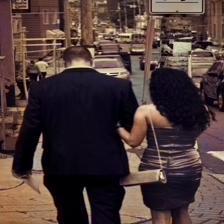}
        \includegraphics[width=0.085\linewidth]{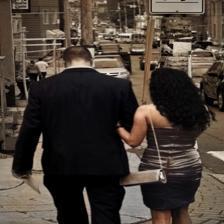}
        \includegraphics[width=0.085\linewidth]{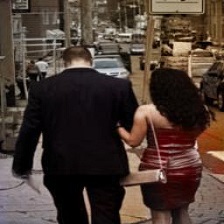}
        \includegraphics[width=0.085\linewidth]{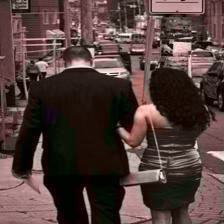}
        \includegraphics[width=0.085\linewidth]{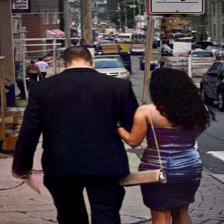}
        \includegraphics[width=0.085\linewidth]{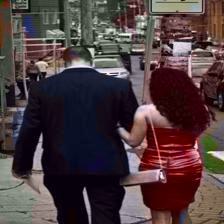}
        \includegraphics[width=0.085\linewidth]{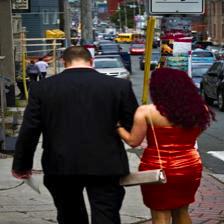}
        \\
 \vspace{1mm}   
        \includegraphics[width=0.085\linewidth]{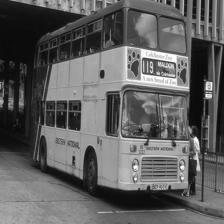}
        \includegraphics[width=0.085\linewidth]{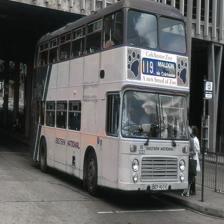}
        \includegraphics[width=0.085\linewidth]{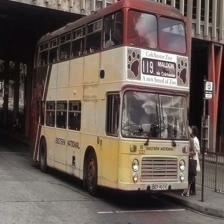}
        \includegraphics[width=0.085\linewidth]{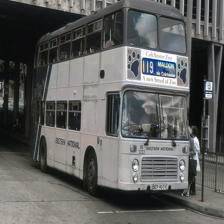}
        \includegraphics[width=0.085\linewidth]{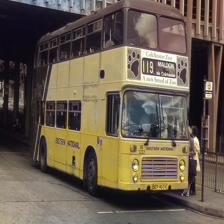}
        \includegraphics[width=0.085\linewidth]{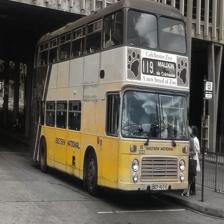}
        \includegraphics[width=0.085\linewidth]{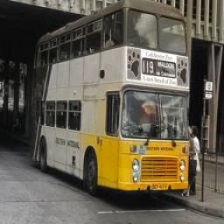}
        \includegraphics[width=0.085\linewidth]{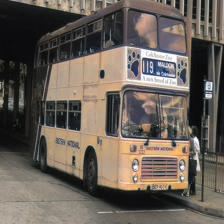}
        \includegraphics[width=0.085\linewidth]{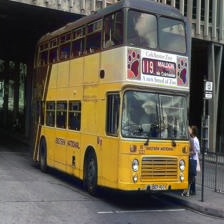}
        \includegraphics[width=0.085\linewidth]{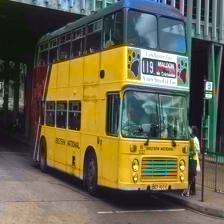}
        \includegraphics[width=0.085\linewidth]{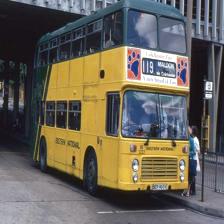}
    \\
\vspace{1mm}    
        \includegraphics[width=0.085\linewidth]{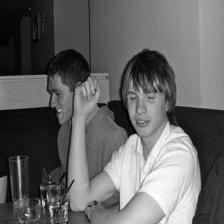}
        \includegraphics[width=0.085\linewidth]{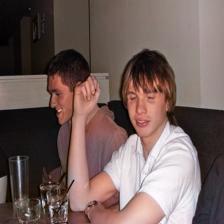}
        \includegraphics[width=0.085\linewidth]{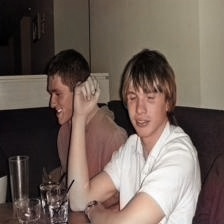}
        \includegraphics[width=0.085\linewidth]{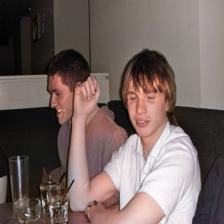}
        \includegraphics[width=0.085\linewidth]{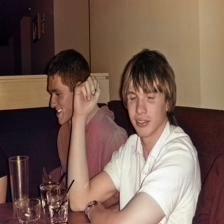}
        \includegraphics[width=0.085\linewidth]{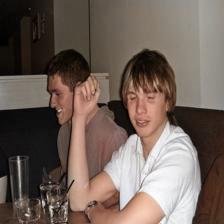}
        \includegraphics[width=0.085\linewidth]{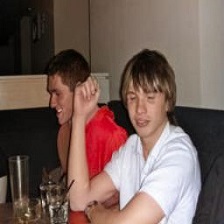}
        \includegraphics[width=0.085\linewidth]{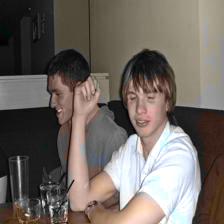}
        \includegraphics[width=0.085\linewidth]{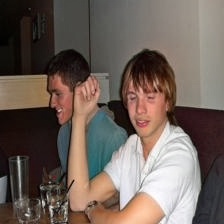}
        \includegraphics[width=0.085\linewidth]{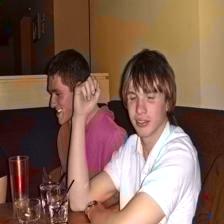}
        \includegraphics[width=0.085\linewidth]{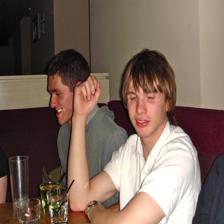}
    \\
\vspace{1mm}
        \includegraphics[width=0.085\linewidth]{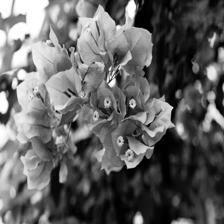}
        \includegraphics[width=0.085\linewidth]{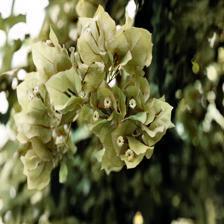}
        \includegraphics[width=0.085\linewidth]{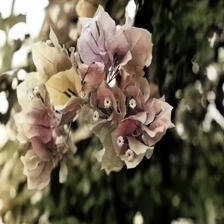}
        \includegraphics[width=0.085\linewidth]{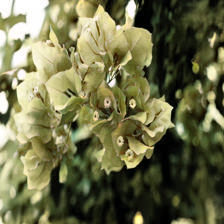}
        \includegraphics[width=0.085\linewidth]{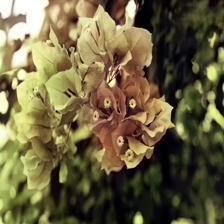}
        \includegraphics[width=0.085\linewidth]{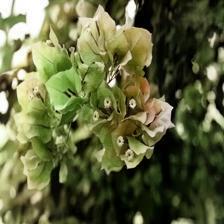}
        \includegraphics[width=0.085\linewidth]{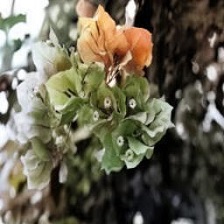}
        \includegraphics[width=0.085\linewidth]{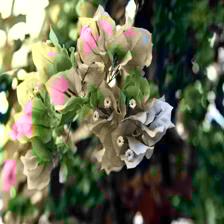}
        \includegraphics[width=0.085\linewidth]{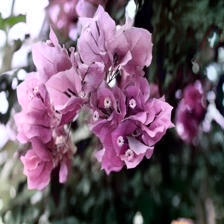}
        \includegraphics[width=0.085\linewidth]{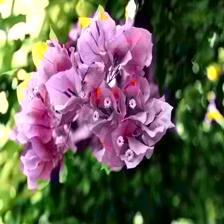}
        \includegraphics[width=0.085\linewidth]{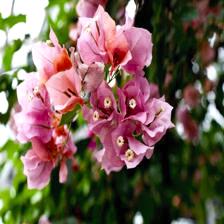}
    \\
\vspace{1mm}
        \includegraphics[width=0.085\linewidth]{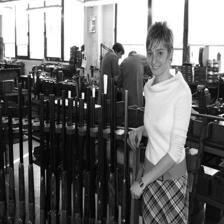}
        \includegraphics[width=0.085\linewidth]{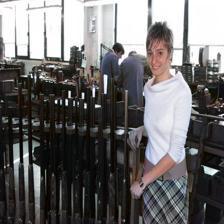}
        \includegraphics[width=0.085\linewidth]{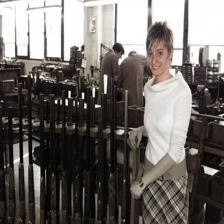}
        \includegraphics[width=0.085\linewidth]{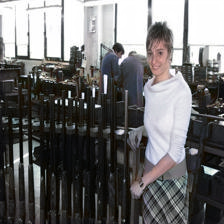}
        \includegraphics[width=0.085\linewidth]{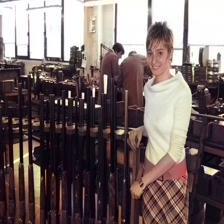}
        \includegraphics[width=0.085\linewidth]{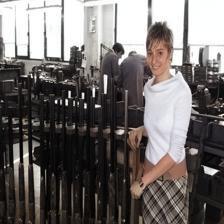}
        \includegraphics[width=0.085\linewidth]{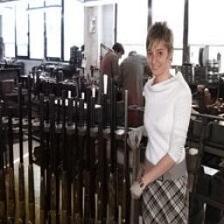}
        \includegraphics[width=0.085\linewidth]{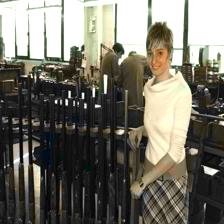}
        \includegraphics[width=0.085\linewidth]{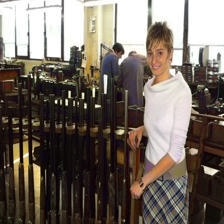}
        \includegraphics[width=0.085\linewidth]{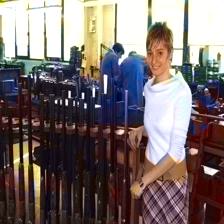}
        \includegraphics[width=0.085\linewidth]{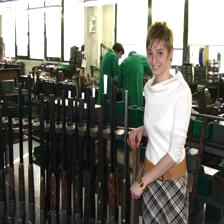}
    \\
\vspace{1mm}
        \includegraphics[width=0.085\linewidth]{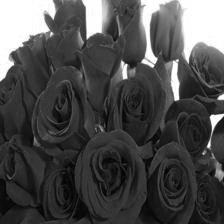}
        \includegraphics[width=0.085\linewidth]{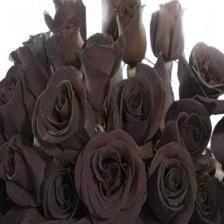}
        \includegraphics[width=0.085\linewidth]{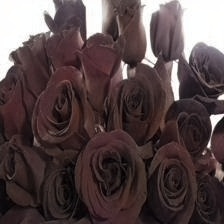}
        \includegraphics[width=0.085\linewidth]{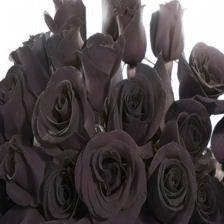}
        \includegraphics[width=0.085\linewidth]{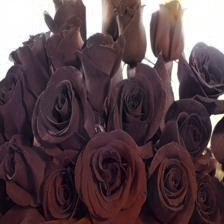}
        \includegraphics[width=0.085\linewidth]{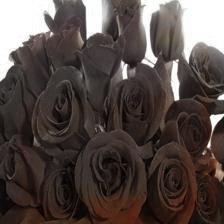}
        \includegraphics[width=0.085\linewidth]{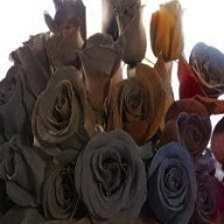}
        \includegraphics[width=0.085\linewidth]{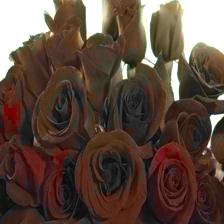}
        \includegraphics[width=0.085\linewidth]{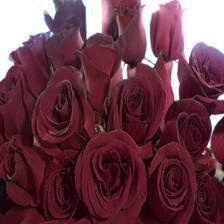}
        \includegraphics[width=0.085\linewidth]{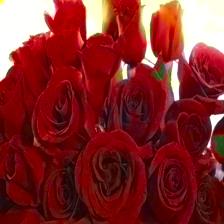}
        \includegraphics[width=0.085\linewidth]{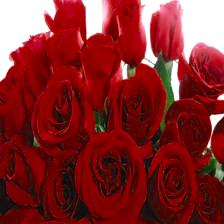}
    \\
\vspace{1mm}
        \includegraphics[width=0.085\linewidth]{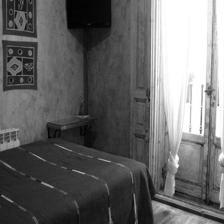}
        \includegraphics[width=0.085\linewidth]{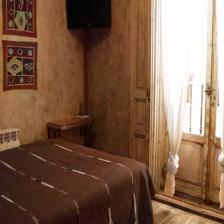}
        \includegraphics[width=0.085\linewidth]{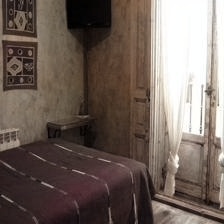}
        \includegraphics[width=0.085\linewidth]{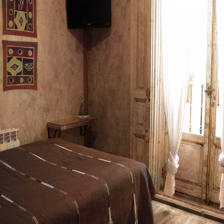}
        \includegraphics[width=0.085\linewidth]{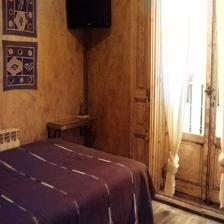}
        \includegraphics[width=0.085\linewidth]{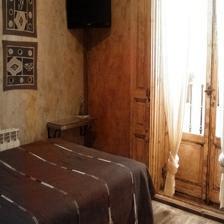}
        \includegraphics[width=0.085\linewidth]{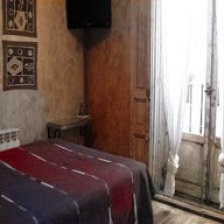}
        \includegraphics[width=0.085\linewidth]{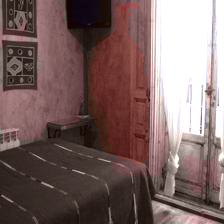}
        \includegraphics[width=0.085\linewidth]{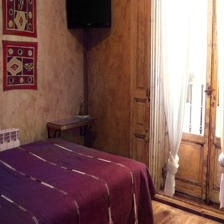}
        \includegraphics[width=0.085\linewidth]{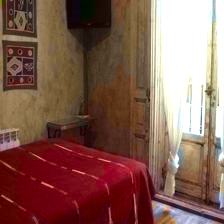}
        \includegraphics[width=0.085\linewidth]{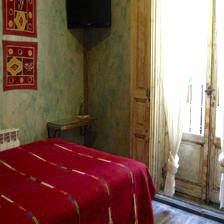}
    \\
\vspace{1mm}
        \includegraphics[width=0.085\linewidth]{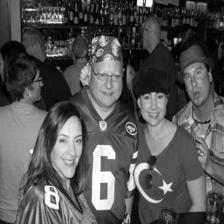}
        \includegraphics[width=0.085\linewidth]{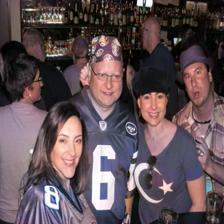}
        \includegraphics[width=0.085\linewidth]{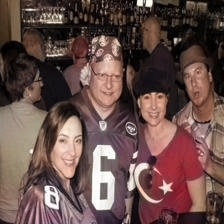}
        \includegraphics[width=0.085\linewidth]{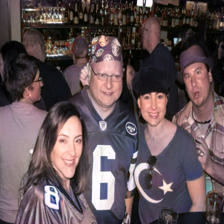}
        \includegraphics[width=0.085\linewidth]{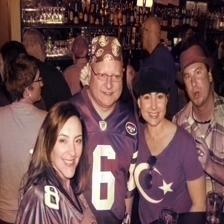}
        \includegraphics[width=0.085\linewidth]{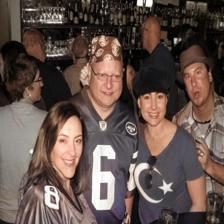}
        \includegraphics[width=0.085\linewidth]{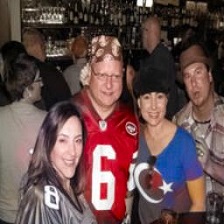}
        \includegraphics[width=0.085\linewidth]{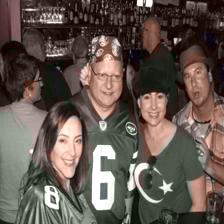}
        \includegraphics[width=0.085\linewidth]{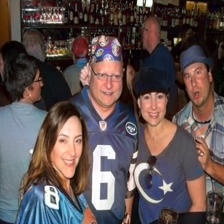}
        \includegraphics[width=0.085\linewidth]{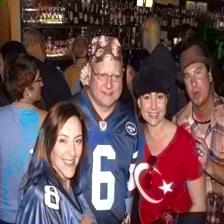}
        \includegraphics[width=0.085\linewidth]{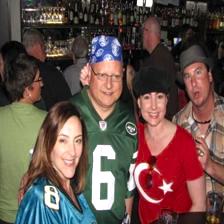} 
        \begin{flushleft}\small
        
             \hspace{2mm} Gray \hspace{2mm} DeOldify\cite{Deoldify}  Iizuka\cite{Iizuka}  Larsson\cite{Larsson} \hspace{1mm} CIC\cite{Zhang_eccv} \hspace{2mm} Zhang\cite{Zhang_tog} \hspace{2mm} Su\cite{Su} \hspace{2mm} Gain\cite{Gain2} \hspace{2mm} DD\cite{DD} \hspace{5mm}  CCC \hspace{5mm} Ground True
             
       \end{flushleft}
    \caption{Some results of our proposed method compared to other state-of-the-art methods.}
    \label{visual}
    \vspace{-4mm}
\end{figure*}

\section{SAM Empowered Object-Selective Color Harmonization}
\label{sec:sam}
As we force regularization of the minor class, there is sometimes a little noise at the object's edge. To make the edge more polished, we proposed SAM-empowered object-selective color harmonization. The SAM is a segmentation model with zero-shot generalization to unfamiliar objects and images without additional training\cite{SAM}. Our proposed algorithm is shown in Algorithm \ref{alg:segmentation}.

\section{Experiment}
\label{experiment}
\textbf{Datasets} We train the proposed model using the \textit{Place365 Train} dataset\cite{Place}.
Our model is developed in a self-supervised manner. We provide no external label for our data during the train. Instead, we generate the model's supervisory signals or labels from the input data during training. 
For testing, we use multiple datasets. We use \textit{Place365 Test} dataset\cite{Place}. The dataset has 328.5k images with 365 scene categories. Besides, we take randomly 50 images from \textit{ImageNet1k Validation}\cite{Imagenet}, \textit{Oxford 102 Flower}\cite{Oxford_flower}, CelebFaces(CelebA)\cite{Celeba}, and COCO\cite{coco}  datasets.\\
\begin{table}[h]
    \centering
    \small
    \caption{Different Hyper-parameter and trade-off factor values for CCC}
    \label{hyperparameter}
    \begin{tabular}{c c c c c c c c c}
        \hline
    H. P.& $\alpha$ & $\beta$ & $\Delta$ & $\Upsilon$ & $\Phi$ &\ $\delta_a$ & $\delta_b$ & $\Psi$  \\
        \hline
    Value & 10 & 100 & 20 & 0.5 & .0046 & 8  & 8 & 500 \\
        \hline
    \end{tabular}
\end{table}
\textbf{Implementation Set Up}
The experiments were conducted on a workstation with an NVIDIA GEFORCE RTX 2080 Ti graphics processing unit (GPU). The network was constructed using PyTorch\cite{pytorch} version 1.28 in Python version 3.10.9. During training, the batch size is set to 64, the Adam optimizer is employed with the learning rate $1 \times 10^{-3}$, and the momentum parameters $\beta$1 = 0.5 and $\beta$2 = 0.999 are used to update and compute the network parameters.  Each ground truth a*b* tensor was resized into $56 \times 56$ size to reduce the complexity of loss calculations. We systematically explore a spectrum of hyper-parameters and trade-off factors for our proposed model, with their values determined through methodical experimental analysis. The corresponding values are detailed in Tab. \ref{hyperparameter}. \\
\begin{table}[h]
    \vspace{-2mm}
    \footnotesize
    \centering
    \caption{Quantitative comparison of our proposed method with the baseline and SOTA methods using the visuals of Fig. \ref{visual}.}
    \label{all}
    \begin{tabular}{p{.80cm}p{.50cm}p{.55cm}p{.58cm}p{.58cm}p{.53cm}p{.45cm}p{.58cm}}
        \hline
         &  MSE$\downarrow$ &  PSNR$\uparrow$  & SSIM$\uparrow$  & LPIPS$\downarrow$ &  UIQI$\uparrow$  &  FID$\downarrow$ &  CNR$\uparrow$  \\
        \hline
         Deoldify & 0.0064 & 21.71 & \textbf{0.898} & 0.183 & 0.872 & 0.395 &  0.615 \\
        \hline
         Iizuka & \textbf{0.0042} & \textbf{22.09} & 0.885 & 0.171 & 0.871 & 0.361 &  0.376 \\
        \hline
         Larsson & 0.0052 & 21.51 & 0.878 & 0.197 & \textbf{0.879} & 0.340 &  0.585 \\
        \hline
         CIC & 0.0061 & 20.82 & 0.864 & 0.194 & 0.864 & 0.319 &  0.619 \\
        \hline
         Zhangs & 0.0054 & 21.78 & 0.878 & \textbf{0.162} & 0.862 & 0.303 &  0.526 \\
        \hline
         Su & 0.0071 & 20.94 & 0.854 & 0.233 & 0.865 & 0.300 &  0.648 \\
        \hline
         Gain & 0.0059 & 21.01 & 0.871 & 0.210 & 0.871 & 0.315 &  0.691 \\
        \hline
         DD & 0.0051 & 21.68 & 0.878 & 0.165 & 0.869 & \textbf{0.276} &  0.798 \\
        \hline
         Our & 0.0055 & 21.46 & 0.878 & 0.207 & 0.872 & 0.288 &  \textbf{0.884} \\
        \hline
    \end{tabular}
\end{table}
\textbf{Evaluation Metrics}
We use mean squared error (MSE\cite{MSE}), peak signal-to-noise ratio (PSNR\cite{PSNR-SSIM}), structural similarity index measure (SSIM\cite{PSNR-SSIM}), learned perceptual image patch similarity (LPIPS\cite{LPIPS}), universal image quality index (UIQI\cite{UIQI}), frechet inception distance score (FID\cite{FID}), and our proposed Chromatic Number Ratio (CNR) to compare our proposed model with the baselines and state-of-the-art (SOTA) colorization methods quantitatively.\\
\textbf{Comparison with Baselines and SOTA:}
We compare our model with eight baselines and SOTA methods: DeOldify\cite{Deoldify}, Iizuka\cite{Iizuka}, Larsson\cite{Larsson}, CIC\cite{Zhang_eccv}, Zhang\cite{Zhang_tog}, Su\cite{Su}, Gain\cite{Gain2}, and DD\cite{DD}. In Fig. \ref{visual}, we compare eight images visually against those methods with gray and ground truth. The figure shows that our proposed CCC method visually outperforms the baselines and SOTA methods. The proposed CCC method effectively colors the minor objects with the majors. 
In Tab. \ref{all}, quantitatively evaluate the images of Fig. \ref{visual}. From the visual and quantitative analysis, we find Deoldify has the best SSIM, Iizuka has the best MSE and PSNR, Larsson has the best UIQI, Zhang has the best LPIPS, DD has the best FID, and CCC has the best CNR. Visually, the CCC has a more plausible major and minor color combination than the others. Therefore, it is evident that MSE, PSNR, SSIM, LPIPS, and FID criteria are not completely suitable for ensuring the presence of minor colors. At first, ensuring the CNR and then maintaining those criteria may be the best possible solution for the appearance of major and minor colors in the generated images.
In Tab. \ref{mse-psnr}, we evaluate our proposed model against seven baselines and SOTA methods across three datasets using regression criteria. The table shows that our method performs well in all datasets and outperforms others in the `Oxford Flower' dataset. Because `ADE' predominantly features natural images, while `Celeba' focuses on human faces, typically presenting a more limited range of color combinations. In contrast, the `Oxford Flowers' dataset is characterized by its diverse array of flower species, each exhibiting a unique and varied color palette. This diversity in coloration within the `Oxford Flowers' dataset provides a more complex and challenging environment for colorization, highlighting the efficacy of our method in handling a wide range of colors and complexities.
\begin{table}[h]
    \vspace{-2mm}
    \small
    \centering
    \caption{Regression loss comparison of our proposed method with the baseline and SOTA methods using multiple datasets.}
    \label{mse-psnr}
    \begin{tabular}{p{1cm}p{.70cm}p{.70cm}p{.70cm}p{.70cm}p{.70cm}p{.70cm}p{.70cm}}
        \hline
        & \multicolumn{2}{c}  {ADE}  & \multicolumn{2}{c}  {Celeba} & \multicolumn{2}{c}  {Ox Flower}  \\
         & \small MSE$\downarrow$  & \small PSNR$\uparrow$ & \small MSE$\downarrow$ & \small PSNR$\uparrow$ & \small MSE$\downarrow$ & \small PSNR$\uparrow$ \\
        \hline
        DeOldify & .0043 & 25.66 & .0045  & 26.06 & .0295 & 16.46\\
        \hline
        Iizuka & \textbf{.0035} & \textbf{26.22} & .0045 & 26.00 & .0211 & 18.01\\
        \hline
        Larsson & .0037 & 25.94 & .0058 & 26.66 & .0245 & 16.85 \\
        \hline
        CIC & .0053 & 24.33 & .0056 & 24.79 & .0261 & 17.16\\
        \hline
        Zhang & .0036 & 26.07 & \textbf{.0041} & \textbf{26.78} & .0295 & 16.80\\
        \hline
        Su & .0038 & 25.37 & .0046 & 25.70 & .0265 & 16.81\\
        \hline
        DD & .0039 & 25.22 & .0066 & 25.70 & .0273 & 16.88\\
        \hline
        CCC & .0058 & 24.03 & .0061 & 24.12 & \textbf{.0201} & \textbf{18.06}\\
        \hline
    \end{tabular}
    \vspace{-2mm}
\end{table}
In Tab. \ref{ssim-uiqi}, we evaluate our proposed model against those methods across three datasets using similarity measurement criteria. The table shows that our method performs well in all datasets while maintaining the minor color structure. Usually, it is easier to achieve good similarity by ignoring the minor color features and focusing only on major ones. However, our proposed method maintains satisfactory similarity while ensuring the minor color features.
\begin{table}[h]
    \vspace{-2mm}
    \small
    \centering
    \caption{Structural Similarity comparison of our proposed method with the baseline and SOTA methods using multiple datasets.}
    \label{ssim-uiqi}
    \begin{tabular}{p{1cm}p{.70cm}p{.70cm}p{.70cm}p{.70cm}p{.70cm}p{.70cm}p{.70cm}}
        \hline
        & \multicolumn{2}{c}  {ADE}  & \multicolumn{2}{c}  {Celeba} & \multicolumn{2}{c}  {Ox Flower}  \\
         & \small SSIM$\uparrow$ & \small UIQI$\uparrow$ & \small SSIM$\uparrow$ & \small UIQI$\uparrow$ & \small SSIM$\uparrow$ & \small UIQI$\uparrow$ \\
        \hline
        DeOldify & \textbf{0.96} & \textbf{0.96} & 0.94  & \textbf{0.94} & \textbf{0.82} & 0.81\\
        \hline
        Iizuka & 0.95 & \textbf{0.96} & \textbf{0.95} & \textbf{0.94} & 0.80 & 0.82\\
        \hline
        Larsson & 0.95 & \textbf{0.96} & 0.94 & 0.93 & \textbf{0.82} & \textbf{0.83}\\
        \hline
        CIC & 0.95 & 0.95 & 0.93 & 0.92 & 0.81 & 0.80\\
        \hline
        Zhang & \textbf{0.96} & \textbf{0.96} & \textbf{0.95} & 0.93 & 0.81 & 0.81\\
        \hline
        Su & 0.92 & \textbf{0.96} & 0.93 & 0.93 & 0.77 & 0.81\\
        \hline
        DD & \textbf{0.96} & \textbf{0.96} & 0.93 & 0.92 & 0.81 & 0.80\\
        \hline
        CCC & 0.91 & 0.95 & 0.92 & 0.92 & 0.80 & 0.80\\
        \hline
    \end{tabular}
    \vspace{-2mm}
\end{table}
In Tab. \ref{lpips-fid}, we evaluate our proposed model against those methods across three datasets using LPIPS and FID criteria. The table shows that our method performs well in all datasets and outperforms others in the `Oxford Flower' dataset. Because `Oxford flowers' have the highest diversity compared to `ADE' and `Celeba.'
\begin{table}[h]
    \vspace{-2mm}
    \small
    \centering
    \caption{Perceptual Image Patch Similarity and frethed image distance comparison of our proposed method with the baseline and SOTA methods using multiple datasets.}
    \label{lpips-fid}
    \begin{tabular}{p{1cm}p{.70cm}p{.70cm}p{.70cm}p{.70cm}p{.70cm}p{.70cm}p{.70cm}}
        \hline
        & \multicolumn{2}{c}  {ADE}  & \multicolumn{2}{c}  {Celeba} & \multicolumn{2}{c}  {Ox. Flower}  \\
         & \small LPIPS$\downarrow$ & \small FID$\downarrow$ & \small LPIPS$\downarrow$ & \small FID$\downarrow$ & \small LPIPS$\downarrow$ & \small FID$\downarrow$ \\
        \hline
        DeOldify & 0.15 & 0.48 & \textbf{0.13}  & 0.43 & 0.35 & 3.85\\
        \hline
        Iizuka & 0.16 & 1.05 & 0.16 & 0.45 & 0.31 & 3.57\\
        \hline
        Larsson & 0.16 & 0.62 & 0.14 & 0.37 & 0.34 & 2.42 \\
        \hline
        CIC & 0.18 & 1.31 & 0.17 & 0.58 & 0.35 & 4.20\\
        \hline
        Zhang & \textbf{0.14} & 1.12 & \textbf{0.13} & 0.49 & 0.34 & 4.72\\
        \hline
        Su & 0.21 & 1.24 & 0.18 & 0.28 & 0.41 & 4.51\\
        \hline
        DD & 0.16 & \textbf{0.30} & 0.16 & \textbf{0.18} & 0.32 & 1.54\\
        \hline
        CCC & 0.15 & 0.90 & \textbf{0.13} & 0.43 & \textbf{0.30} & \textbf{1.51}\\
        \hline
    \end{tabular}
    \vspace{-2mm}
\end{table}
In Tab. \ref{csdm}, we evaluate our proposed model against those methods across five datasets using CNR criteria. The table shows that our method outperforms all methods in all datasets. The main objective of our proposed model is to ensure the presence of minor colors along with major. Minor color confirmation makes color images more diverse because an image contains one or two major colors as well as more minor colors.
\begin{table}[h]
    \vspace{-2mm}
    \small
    \centering
    \caption{CNR comparison of our proposed method with the baseline and SOTA methods using multiple datasets.}
    \label{csdm}
    \begin{tabular}{p{1cm}p{.70cm}p{.70cm}p{.70cm}p{1.5cm}p{1cm}}
        \hline
         & \small ADE & \small Celeba & \small COCO & \small Ox. Flower & \small ImageNet  \\
        \hline
        DeOldify & 0.77 & 0.62 & 1.43  & 0.69 & 0.61 \\
        \hline
        Iizuka & 0.78 & 0.51 & 1.49 & 0.58 & 0.49 \\
        \hline
        Larsson & 0.77 & 0.64 & 0.73 & 0.73 & 0.63 \\
        \hline
        CIC & 0.81 & 0.86 & 1.57 & 0.67 & 0.58 \\
        \hline
        Zhang & 0.73 & 0.66 & 1.05 & 0.66 & 0.49 \\
        \hline
        Su & 0.77 & 0.80 & 2.89 & 0.66 &0.66 \\
        \hline
        DD & 1.25 & 1.07 & 2.23 & 0.88 & 0.94 \\
        \hline
        CCC & \textbf{1.90} & \textbf{1.33} & \textbf{3.53} & \textbf{0.96} & \textbf{1.13} \\
        \hline
    \end{tabular}
    \vspace{-2mm}
\end{table}

\section{Conclusion}
\label{conclusion}
Automatic colorization of grayscale photographs with objects of varying colors and sizes is complex due to inter- and intra-object color variations and the limited area occupied by principal items. The learning process often favors dominating features, leading to biased models. A weighted function can address feature imbalance, assigning greater importance to minority features. In this paper, we propose a set of formulas to convert color values into corresponding color classes and vice versa. To achieve optimal performance, we optimize the class levels and establish a trade-off between the weights of major and minor classes, considering both types of classes for accurate class prediction. We also propose SAM-empowered object selective color harmonization that improves the stability of minor classes. We propose a novel color picture assessment measure called Chromatic Number Ratio (CNR) to assess color component richness quantitatively. We evaluated our model against eight baseline and SOTA models using five datasets, and experimental findings show that the proposed model surpasses previous models in terms of visualization and CNR measurement criteria while maintaining satisfactory performance in other regression criteria, MSE, PSNR, similarity criteria SSIM, LPIPS, UIQI, and generative criteria FID.

\end{document}